\documentclass[conference]{IEEEtran}
\IEEEoverridecommandlockouts
\usepackage{cite}
\usepackage{amsmath,amssymb,amsfonts}
\usepackage{algorithmic}
\usepackage{algorithm}
\usepackage{graphicx}
\usepackage{textcomp}
\usepackage{xcolor}
\def\BibTeX{{\rm B\kern-.05em{\sc i\kern-.025em b}\kern-.08em
    T\kern-.1667em\lower.7ex\hbox{E}\kern-.125emX}}
\usepackage{subfigure}

\newcommand{\etal}{\textit{et al}.}
\newcommand{\ie}{\textit{i}.\textit{e}.}
\newcommand{\eg}{\textit{e}.\textit{g}.}

\begin{document}

\title{Single-Net Continual Learning with \\ Progressive Segmented Training
}

\author{\IEEEauthorblockN{Xiaocong Du}
\IEEEauthorblockA{\textit{School of ECEE} \\
\textit{Arizona State University}\\
Tempe, AZ, USA \\
xiaocong@asu.edu}\\
\and
\IEEEauthorblockN{Gouranga~Charan}
\IEEEauthorblockA{\textit{School of ECEE} \\
\textit{Arizona State University}\\
Tempe, AZ, USA \\
gcharan@asu.edu}
\and
\IEEEauthorblockN{Frank Liu}
\IEEEauthorblockA{\textit{CSMD}\\ 
\textit{Oak Ridge National Lab}\\
Oak Ridge, TN, USA \\
liufy@ornl.gov}
\and
\IEEEauthorblockN{Yu Cao}
\IEEEauthorblockA{\textit{School of ECEE} \\
\textit{Arizona State University}\\
Tempe, AZ, USA \\
ycao@asu.edu}
}



\maketitle

\begin{abstract}
There is an increasing need of continual learning in dynamic systems, such as the self-driving vehicle, the surveillance drone, and the robotic system. Such a system requires learning from the data stream, training the model to preserve previous information and adapt to a new task, and generating a single-headed vector for future inference. Different from previous approaches with dynamic structures, this work focuses on a single network and model segmentation to prevent catastrophic forgetting. Leveraging the redundant capacity of a single network, model parameters for each task are separated into two groups: one important group which is frozen to preserve current knowledge, and secondary group to be saved (not pruned) for a future learning. A fixed-size memory containing a small amount of previously seen data is further adopted to assist the training. Without additional regularization, the simple yet effective approach of Progressive Segmented Training (PST) successfully incorporates multiple tasks and achieves the state-of-the-art accuracy in the single-head evaluation on CIFAR-10 and CIFAR-100 datasets. Moreover, the segmented training significantly improves computation efficiency in continual learning at the edge.
\end{abstract}

\begin{IEEEkeywords}
Continual learning, convolutional neural network, deep learning, computer vision, image recognition
\end{IEEEkeywords}

\section{Introduction}\label{sec:intro}
The rapid advancement of computing and sensing technology has enabled many new applications, such as the self-driving vehicle, the surveillance drone, and the robotic system. Compared to conventional edge devices (\eg~cell phone or smart home devices), these emerging devices are required to deal with much more complicated, dynamic situations. One of the necessary attributes is the capability of continual learning: when encountering a sequence of tasks over time, the learning system should capture the new observation and update its knowledge (\ie~the network parameters~\cite{kirkpatrick2017overcoming,zenke_continual_2017}) in real time, without interfering or overwriting previously acquired knowledge. 
Recent literature~\cite{aljundi_online_2019,chaudhry_riemannian_2018,chaudhry_efficient_2019,  lee2017overcoming,lopez-paz_gradient_2017,rebuffi_icarl_2016,zhang_regularize_2019,zhang_class-incremental_2019} have intensively studied this topic, and it is believed that, in order to learn a data stream continually, such a system should have the following features:

\textbf{Online adaption.} The system should be able to update its knowledge according to a continuum of data, without independent and identically distributed (i.i.d.) assumption on this data stream. For a dynamic system (\eg~a self-driving vehicle), it is preferred that such adaption is completed locally and in real time.

\textbf{Preservation of prior knowledge.} When new data arrives in a stream, previous data are very limited or even no longer exists. Yet the acquired knowledge from previous data should not be \textit{forgotten} (\ie~overwritten or deteriorated due to the learning of new data). In other words, the prior distribution of the model parameters should be preserved.

\textbf{Single-head evaluation.} The network should be able to differentiate the tasks and achieve successful inter-task classification without the prior knowledge of the task identifier (\ie~which task current data belongs to).  In the case of single-head, the network output should consist of all the classes seen so far. In contrast, multi-head evaluation only deals with intra-task classification where the network output only consists of a subset of all the classes. Multi-head classification is more appropriate for multi-task learning than continual learning~\cite{aljundi_online_2019}.

\textbf{Resource constraint.} The resource usage such as the model size, the computation cost, and storage requirements should be bounded during continual learning from sequential tasks, rather than increasing proportionally or even exponentially over time.

For the aforementioned features, one of the serious challenges is \textit{catastrophic forgetting} of the prior knowledge.  McCloskey \etal~\cite{mccloskey1989catastrophic} first identified the catastrophic forgetting problem in the connectionist networks. Henceforth, various solutions to mitigate catastrophic forgetting have been proposed. These solutions can be categorized into two families: 
\textbf{(1) Dynamic network structure}. These methods \cite{rusu_progressive_2016, xu_reinforced_nodate, yoon_lifelong_2017, serra_overcoming_2018,zhang_regularize_2019,zhang_class-incremental_2019} usually expand the new knowledge by growing the network structure. For example, \cite{rusu_progressive_2016} progressively adds new network branches for new tasks and keeps previously learned features in lateral connections. In this case, prior knowledge and new knowledge are usually separated into different feedforward paths. Moreover, the newly added branches have never been exposed to the previous data and thus is blind to previous tasks. Due to these fundamental reasons, the performance of dynamic architectures on the single-head classification lags behind, although they were able to maintain the accuracy in multi-head classification with the priori of task identification. The second family is \textbf{(2) single network structure}. In contrast to a dynamic structure, these methods learn sequential tasks with a single, static network structure all the time. The knowledge of prior and new tasks are packed in a single network that is exposed to all tasks over time. In this case, the challenge is shifted to minimizing the interference among tasks and preserving prior knowledge in the same network. As a contemporary neural network has a large capacity to accommodate multiple tasks,  we believe a single network provides a promising basis for continual learning.

In the family of the single-network methods, previous works have explored the regularization methods~\cite{ferrari_memory_2018,chaudhry_riemannian_2018,kirkpatrick_overcoming_2016,li_learning_2018,zenke_continual_2017}, the parameter isolation methods~\cite{ferrari_piggyback_2018,mallya_packnet:_2018} and the memory rehearsal methods~\cite{castro2018end,chaudhry_efficient_2019,javed2018revisiting,lopez-paz_gradient_2017,rebuffi_icarl_2016}. The regularization methods leverage a penalty term in the loss function to regularize the parameters when updating for new tasks. However, as more and more tasks appear, the parameters tend to be biased towards the new tasks, and the system gradually drifts away from previous distribution. To mitigate such a knowledge asymmetry, regularization methods can be combined with memory rehearsal methods~\cite{farquhar2018towards,nguyen2017variational}. Recent works such as iCaRL~\cite{rebuffi_icarl_2016} and GEM~\cite{lopez-paz_gradient_2017} have proven the efficacy of replaying the memory (\ie~train the system with a subset of the previously seen data) in abating the network parameters drifting far away from previous knowledge. Parameter isolation approaches~\cite{ferrari_piggyback_2018,mallya_packnet:_2018} allocate subsets of parameters for previous tasks and prune the rest for learning new tasks. In this case, the rest of the parameters no longer contain prior knowledge, violating the aforementioned properties of an ideal continual learning system. For instance, PackNet~\cite{ferrari_piggyback_2018} and Piggyback~\cite{mallya_packnet:_2018} achieve strong performance on multi-head evaluation but not on single-head.

\begin{figure*}[!t]
\begin{center}
\includegraphics[width=\textwidth]{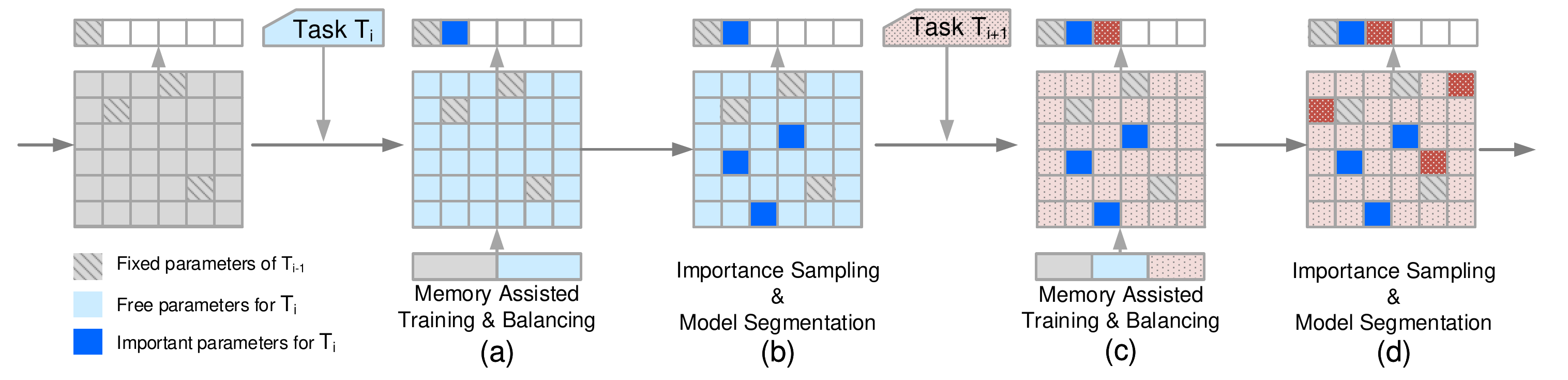}
\end{center}
\caption{The flow chart of Progressive Segmented Training (PST). (a) We allow the current task $T_i$ and a memory set to update the free parameters $\Theta_{free}$ (in light blue) in the network while sharing fixed parameters $\Theta_{fixed}$ (in grey) learned from previous tasks.
The fixed-size memory set is used to keep the balance of training among various tasks.
(b) We sort and select important parameters $\Theta_{important}$ (in dark blue) for task $T_i$, and reinforce them by retraining. These important parameters are kept frozen and will not be updated by future tasks. Different from ~\cite{mallya_packnet:_2018} and~\cite{ferrari_piggyback_2018}, the secondary parameters (in light blue) are NOT pruned in PST. Instead, new tasks will start from secondary parameters and update the network, which is essential to achieve single-head classification. For a new task $T_{i+1}$, the above training routine repeats in (c) and (d), so on and so forth.}
\label{fig:flow}
\end{figure*}

To achieve continual learning with the preservation of prior knowledge, we propose single-net continual learning with Progressive Segmented Training, namely PST, as shown in Fig.~\ref{fig:flow}. When new data comes in, PST adapts the network parameters with memory-assisted balancing, then important parameters are identified according to their contribution to this task. Next, to alleviate catastrophic forgetting, PST performs model segmentation by reinforcing important parameters (through retraining) and then freezing them in the future training; while the secondary parameters will be saved (not pruned) and updated by future tasks. 
Through experiments on CIFAR-10~\cite{krizhevsky2009learning} and CIFAR-100~\cite{krizhevsky2009learning} dataset with modern deep neural networks, we demonstrate that PST achieves state-of-the-art single-head accuracy and successfully preserves the previously acquired knowledge in the scenario of continual learning. Moreover, benefiting from model segmentation, the amount of computation needed to learn a new task keeps reducing. This property brings PST high efficiency in computation as compared to other regularization methods.

The contribution of this paper is as follows:
\begin{itemize}
\item We summarize important features of a successful continual learning system and propose a novel training scheme, namely Progressive Segmented Training (PST),  to mitigate catastrophic forgetting in continual learning. 

\item Different from previous works in which new observation overwrites the entire acquired knowledge, PST leverages parameter segmentation for each task to prevent knowledge overwriting or deterioration.
Experiments on CIFAR-10 and CIFAR-100 dataset prove that PST successfully alleviates catastrophic forgetting and reaches state-of-the-art single-head accuracy in the learning of streamed data, with 24$\times$ reduction in computation cost as compared to the previous work.

\item We demonstrate a detailed ablation study and discussion to analyze the role of each component in PST. 

\end{itemize}

\begin{algorithm}[!t]
\caption{PST training routine}
\label{algo1}
\textbf{Input}:$\left\{X^{s}, \ldots, X^{t}\right\}$ \\
\textbf{Require} $\Theta = (\Theta_{fixed}; \Theta_{free} )$ \\
\textbf{Require} $\mathcal{P}=(P_{1}, \ldots, P_{s-1})$  
\begin{algorithmic}[1] 
\STATE Memory-assisted training and balancing: $\Theta_{free} \rightarrow \Theta'_{free}$
\STATE Importance sampling: identify $\Theta_{important}$ in $\Theta'_{free} $ 
~\STATE Model segmentation: $\Theta_{important} \rightarrow \Theta'_{important}$ 
\STATE $(\Theta_{fixed};\Theta'_{important}) \rightarrow \Theta'_{fixed}$  
\STATE $\Theta_{secondary} \rightarrow \Theta'_{free} $ 
\end{algorithmic}
\textbf{Output}: $\Theta' = (\Theta'_{fixed};\Theta'_{free})$ \\
\textbf{Output}:  $\mathcal{P'}=(P_{1}, \ldots, P_{t})$
\end{algorithm}

\section{Method}\label{sec:method}
In this section, we first describe the terminology and algorithm of PST. Then we interpret three major components: memory-assisted training and balancing, importance sampling and model segmentation in Section~\ref{sec:training}, Section~\ref{sec:importance} and Section~\ref{sec:segmentation}, respectively.

\subsection{Overview of PST}
\paragraph{Terminology} The continual learning problem can be formulated as follows: the machine learning system is continuously exposed to a stream of labeled input data $X^{1}, X^{2}, \ldots $, where $X^{y}=\left\{x_{1}^{y}, \ldots, x_{n_{y}}^{y}\right\}$ correspond to all examples of class $y \in \mathbb{N}$. When the new task $\left\{X^{s}, \ldots, X^{t}\right\}$ comes in, the data of old tasks $\left\{X^{1}, \ldots, X^{s-1}\right\}$ are no longer available, except a small amount of previously seen data stored in the memory set $\mathcal{P}=\left(P_{1}, \ldots, P_{s-1}\right)$.

For a modern deep neural network such as VGG-Net~\cite{simonyan2014very} and ResNet~\cite{he2016deep}, the network parameter $\Theta$ usually consists of feature extractor $\varphi: \mathcal{X} \rightarrow \mathbb{R}^{d}$ and classification weight vectors $w \in \mathbb{R}^{d}$. The network keeps updating its parameter $\Theta$ according to the previously seen data $\mathcal{X}$, in order to predict labels $\mathcal{Y}^*$ with its output $\mathcal{Y}=w^{\top} \varphi(\mathcal{X})$.
During training the network with data corresponding to classes $\left\{X^{1}, \ldots, X^{s-1}\right\}$, our target is to minimize the loss function $\mathcal{L}(\mathcal{Y}; \mathcal{X}_{s-1}; \Theta)$ of this $(s-1)$-class classifier. Similarly, with the introduction of a new task with classes $\left\{X^{s}, \ldots, X^{t}\right\}$, the target now is to minimize $\mathcal{L}(\mathcal{Y}; \mathcal{X}_t; \Theta)$ of this $t$-class classifier.

\paragraph{Training routine}
Every time when a new task is available, PST calls a training routine (Fig.~\ref{fig:flow} and Algorithm~\ref{algo1}) to update the parameter $\Theta$ to $\Theta'$, and the memory set $\mathcal{P}$ to $\mathcal{P'}$, according to the current training data $\left\{X^{s}, \ldots, X^{t}\right\}$ and a small amount of previously seen data (memory set) $\mathcal{P}$. The training routine consists of three major components: (1) memory-assisted training and balancing, (2) importance sampling and (3) model segmentation, as illustrated in the following subsections.

\subsection{Memory-assisted training and balancing}\label{sec:training}
Fig.~\ref{fig:flow} illustrates PST training routine for task $T_i$ and task $T_{i+1}$. In Fig.~\ref{fig:flow}a, which is the moment that task $T_i$ comes in, the network consists of two portions: parameters $\Theta_{fixed}$ (grey blocks) are fixed for previous tasks, and parameters $\Theta_{free}$ (light blue blocks) are trainable for current and future tasks. We allow $\Theta_{free}$ to be updated for task $T_i$, with $\Theta_{fixed}$ included in the feedforward path. 
To mitigate the parameters bias towards new task, a memory set is used to assist the training. The memory set is sampled uniformly and randomly from all the classes in previous tasks, which is a simple yet highly efficient approach, as explained in RWalk work~\cite{chaudhry_riemannian_2018}. For example, if the memory budget is $K$ and $s-1$ classes have been learned in previous tasks, then the memory set stores $\frac{K}{s-1}$ images for each class.
We mix samples from this memory set with equal samples per class from the current task, \ie~$K$ samples of the memory and $\frac{K}{s-1}\times (t-s+1)$ samples from current task, and provide them to the network: 
\textit{(\romannumeral1)} for a few epochs at the beginning of the training;
\textit{(\romannumeral2)} periodically (\eg~every 3 epochs) during training;
\textit{(\romannumeral3)} for a few epochs at the end of the training to fine-tune classification layer (\textit{\romannumeral1, \romannumeral2, \romannumeral3}~are noted in Fig.~\ref{fig:LC}).  In comparison to most related works, which adopt the single-stage (step {\romannumeral2}) optimization technique, the proposed three-step optimization strategy performs much better. One of the primary reasons behind catastrophic forgetting is knowledge drift in both feature extraction and classification layers. The three-pronged strategy helps minimize this drift in the following ways:  step \textit{\romannumeral1} provides a well-balanced initialization; step \textit{\romannumeral2} reviews previous data and thus, consolidates previous learned knowledge for the entire network; step \textit{\romannumeral3} corrects bias by balancing classification layers, which is simple yet efficient as compared to \cite{abs-1905-13260} that utilizes an extra Bias Correction Layer after the classifier. 
After memory-assisted training and balancing, the network parameters are updated from $\Theta =(\Theta_{fixed}; \Theta_{free})$ to $\Theta'=(\Theta_{fixed}; \Theta'_{free})$, as stated in Algorithm 1 line 1.

\subsection{Importance sampling}\label{sec:importance}
After the network has learned on task $T_i$, PST samples crucial learning units for the current task: for feature extraction layers (\ie~convolutional layers), PST samples important \textit{filters}; for fully-connected layers, PST samples important \textit{neurons}. The definitions of \textit{filter} and \textit{neuron} are as follows:

The $l$-th convolutional layer can be formulated as: the output of this layer $\mathcal{Y}_l = \mathcal{X}_l\ast\Theta_l$, where $\Theta_l \in \mathbb{R}^{O_l\times I_l\times K\times K}$. The set of weights that generates the $o$-th output feature map is denoted as a \textit{filter} $\Theta^o_l$, where $\Theta^o_l\in\mathbb{R}^{I_l\times K\times K}$. 
The $l$-th fully-connected layer can be represented by: $\mathcal{Y}_l = \mathcal{X}_l \cdot \Theta_l$, where $\Theta_l \in \mathbb{R}^{O_l \times I_l}$. The set of weights $\Theta_l^t$ that connected to the $t$-th class can be denoted as a \textit{neuron}, where $\Theta_l^t \in \mathbb{R}^{1\times I_l}$. 

The filter/neuron sampling is based on an importance score that is adopted in PST to measure the effect of a single filter/neuron on the loss function, \ie~the importance of each filter/neuron. The importance score is developed from the Taylor Expansion of the loss function. Previously, Molchanov \etal ~\cite{molchanov2016pruning} applied it on pruning secondary parameters.
The importance score represents the difference between the loss with and without each filter/neuron. In other words, if the removal of a filter/neuron leads to relatively small accuracy degradation, this unit is recognized as an unimportant unit, and vice versa. Thus, the objective function to get the filter with the highest importance score is formulated as:
\begin{equation}
 \resizebox{0.95\columnwidth}{!}{$
 \underset{\Theta_l^o} {argmin} {|\Delta\mathcal{L}(\Theta_l^o)|} \Leftrightarrow \underset{\Theta_l^o} {argmin} |\mathcal{L}(\mathcal{Y};\mathcal{X};\Theta) -\mathcal{L}(\mathcal{Y};\mathcal{X};\Theta_l^o=\mathbf{0})|
$}
\end{equation} 
Using the first-order of Taylor Expansion of $|\mathcal{L}(\mathcal{Y};\mathcal{X};\Theta)-\mathcal{L}(\mathcal{Y};\mathcal{X};\Theta_l^o=\mathbf{0})|$ at $\Theta_l^o = \mathbf{0}$, we get:
\begin{equation}\label{math:filter_score}   
 \resizebox{0.950\linewidth}{!}{$
 |\Delta\mathcal{L}(\Theta_l^o)| \simeq |\frac{\partial{\mathcal{L}}(\mathcal{Y};\mathcal{X};\Theta)}{\partial{\Theta_l^o}}\Theta_l^o|  
    =\sum_{i=0}^{I_l}\sum_{m=0}^{K}\sum_{n=0}^{K}|\frac{\partial{\mathcal{L}}(\mathcal{Y};\mathcal{X};\Theta)}{\partial{\Theta_l^{o, i, m, n}}} \Theta_l^{o, i, m, n}| 
$}
\end{equation} 
where $\frac{\partial{\mathcal{L}}(\mathcal{Y};\mathcal{X};\Theta)}{\partial{\Theta_l^{o, i, m, n}}} $ is the gradient of the loss function with respect to parameter $\Theta_l^{o, i, m, n}$.

Similarly, the saliency score of a neuron is derived as:
\begin{equation} \label{math:neuron_score}  
 \resizebox{0.80\columnwidth}{!}{$
   |\Delta\mathcal{L}(\Theta_l^t)| \simeq |\frac{\partial{\mathcal{L}}(\mathcal{Y};\mathcal{X};\Theta)}{\partial{\Theta_{l}^t}}\Theta_{l}^t|   = \sum_{i=0}^{I_l}|\frac{\partial{\mathcal{L}}(\mathcal{Y};\mathcal{X}; \Theta)}{\partial{\Theta_l^{t, i}}} \Theta_l^{t, i}| 
$}
\end{equation} 
where $\frac{\partial{\mathcal{L}}(\mathcal{Y};\mathcal{X}; \Theta)}{\partial{\Theta_l^{t, i}}} $ is the gradient of the loss with respect to parameter $\Theta_l^{t, i}$.

Based on the importance score, we sort the learning units layer by layer and identify the top $\beta$ units (dark blue blocks in Fig.~\ref{fig:flow}b). In the following model segmentation step, we deal with the location of important parameters, rather than the value of these parameters, which will be explained in the next subsection.
$\beta$ is an empirical hyper-parameter that should be approximately proportional to the complexity of the current task. For example, when incrementally learning 10 classes of CIFAR-100 at a time, $\beta$ can be 10\%; when learning 20 classes per task, $\beta$ can be 20\%. 

Due to the nature of continual learning, the total number of tasks is not known beforehand, so the network can be reserved with a larger capacity in order to freeze enough knowledge for previous tasks and leave enough space for future tasks. Once the continual learning is complete, one can leverage model compression approaches~\cite{han2015learning,haoli,hubara2017quantized,du2019jetcas,gong2014compressing} to compress the model size. 
It is also worth mentioning that importance sampling is only performed once after each task, so that the computation cost of this step is negligible.

\begin{figure*}[!t]
\begin{center}
\includegraphics[width=\textwidth]{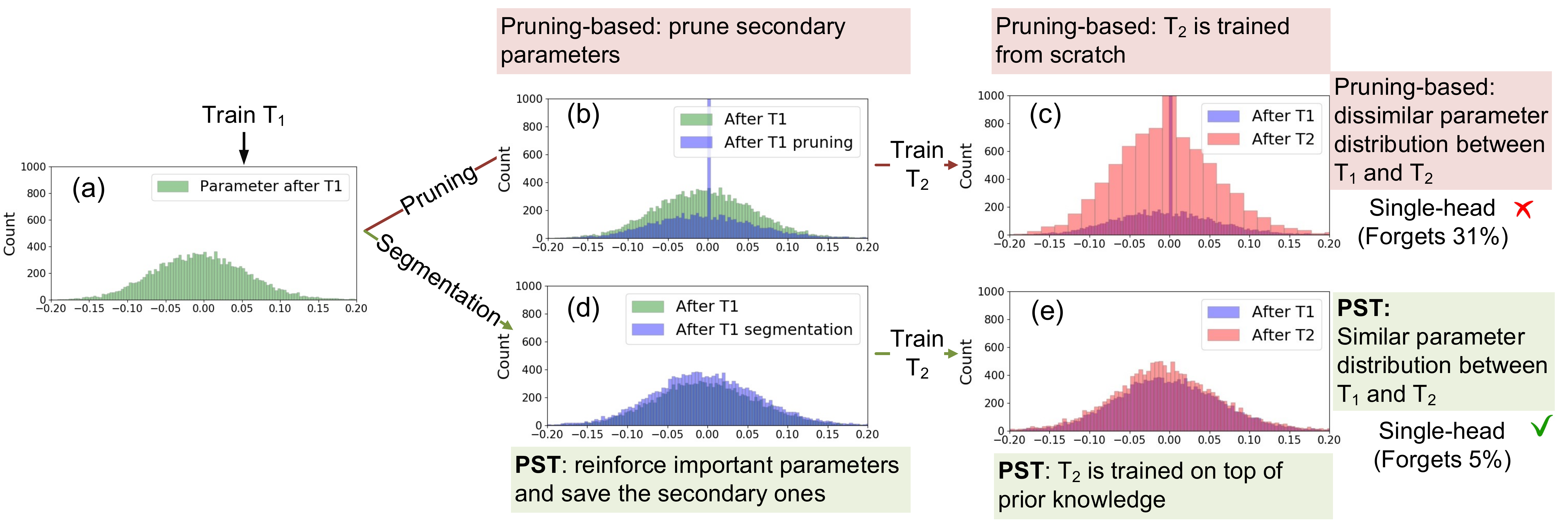}
\end{center}
\caption{Comparison of weight distribution between pruning-based approaches and our PST. Pruning-based approaches lose prior knowledge due to pruning, and PST preserves prior knowledge by segmentation.}
\label{fig:distribution}
\end{figure*}

\subsection{Model segmentation and reinforcement}
\label{sec:segmentation}

After important units are sampled according to the importance score, current network parameter $\Theta' = (\Theta_{fixed};\Theta_{important};\Theta_{secondary})$, where $\Theta_{fixed}$ are the frozen parameters for all the previous tasks, $\Theta_{important}$ are important parameters for the current task, and $\Theta_{secondary}$ are unimportant parameters for the current task, as stated in Algorithm~\ref{algo1} line 2.  Our ideal target is to reinforce $\Theta_{important}$ in a way such that their contribution to the current task is as crucial as possible. Previously, Liu \etal ~\cite{liu2018rethinking} observed that the sampled network architecture itself (rather than the selected parameters) is more indispensable to the learning efficacy. Inspired by this conclusion, we keep the $\Theta_{fixed}$ and $\Theta_{secondary}$ intact, randomly initialize $\Theta_{important}$ and retrain them with current training data assisted by memory set to obtain $\Theta'_{important}$. This step reinforces the contribution of $\Theta_{important}$ to the learning, as proved by our experimental results demonstrated in Fig.~\ref{fig:distribution} and Table~\ref{table:hybrid}. After model segmentation, $\Theta'_{important}$ along with the aforementioned $\Theta_{fixed}$ will be kept frozen in the future tasks, and $\Theta_{secondary}$ will be used to learn new knowledge.


\section{Experimental results}\label{sec:experiment}
In this section, we present experimental results to verify the efficacy of PST. The experiments are performed with PyTorch~\cite{paszke2017automatic} on one NVIDIA GeForce RTX 2080 platform.

\textbf{Datasets.} The CIFAR~\cite{krizhevsky2009learning} dataset consists of 50,000 training images and 10,000 testing images in color with size $32\times32$. There are 10 classes for CIFAR-10 and 100 classes for CIFAR-100. In Section~\ref{sec:cifar10}, CIFAR-10 is divided into 2 tasks, \ie~5 classes per task, to provide a comprehensive analysis of PST. In Section~\ref{sec:cifar100} and \ref{sec:edge}, following iCaRL~\cite{rebuffi_icarl_2016}, CIFAR-100 is divided into 5, 10, 20 or 50 classes per task, to demonstrate extensive experiments. For each experiment, we shuffle the class order and run 5 times to report the average accuracy.

\textbf{Network structures.} In the following experiment, the structure and size of VGG-16~\cite{simonyan2014very} we use for CIFAR-10 dataset follows~\cite{simonyan2014very}. The structure and size of 32-layer ResNet for CIFAR-100 dataset follows the design of iCaRL~\cite{rebuffi_icarl_2016}.  Each convolutional layer in VGG-16 and ResNet is followed by a batch normalization layer~\cite{ioffe2015batch}. As aforementioned in Section~\ref{sec:importance}, the number of classes will occur is unknown in a continual learning scenario. Thus, we leave $1.2\times$ space at the final classification layer in the following experiments, \ie~12 outputs for CIFAR-10 and 120 outputs for CIFAR-100. It is worth mentioning that the number of classes reserved at the final classification layer does not affect the overall performance, as there is no feedback from vacant classes.

\textbf{Experimental setup.} Standard Stochastic Gradient Descent with momentum $0.9$ and weight decay 5E-4 are used for training. The initial learning rate is set to 0.1 and is divided by 10 for every 40\% and 80\% of the total training epochs.  On CIFAR-10 and CIFAR-100 datasets, we train 180 and 100 epochs at the stage of memory-assisted training and balancing, 120 and 60 epochs at the stage of model segmentation. 
$\beta$ is set as proportional to the complexity of the current task. For example, when incrementally learning 5 classes of CIFAR-10 at a time, $\beta$ is set to 50\%; when incrementally learning 10 classes of CIFAR-100 at a time, $\beta$ is set to 10\%. The memory storage is set as $K = 2000$ images for a fair comparison with the previous work~\cite{rebuffi_icarl_2016}.

\begin{figure}[!t]
\centering
\subfigure[Single-head accuracy]{
\includegraphics[width=0.95\columnwidth]{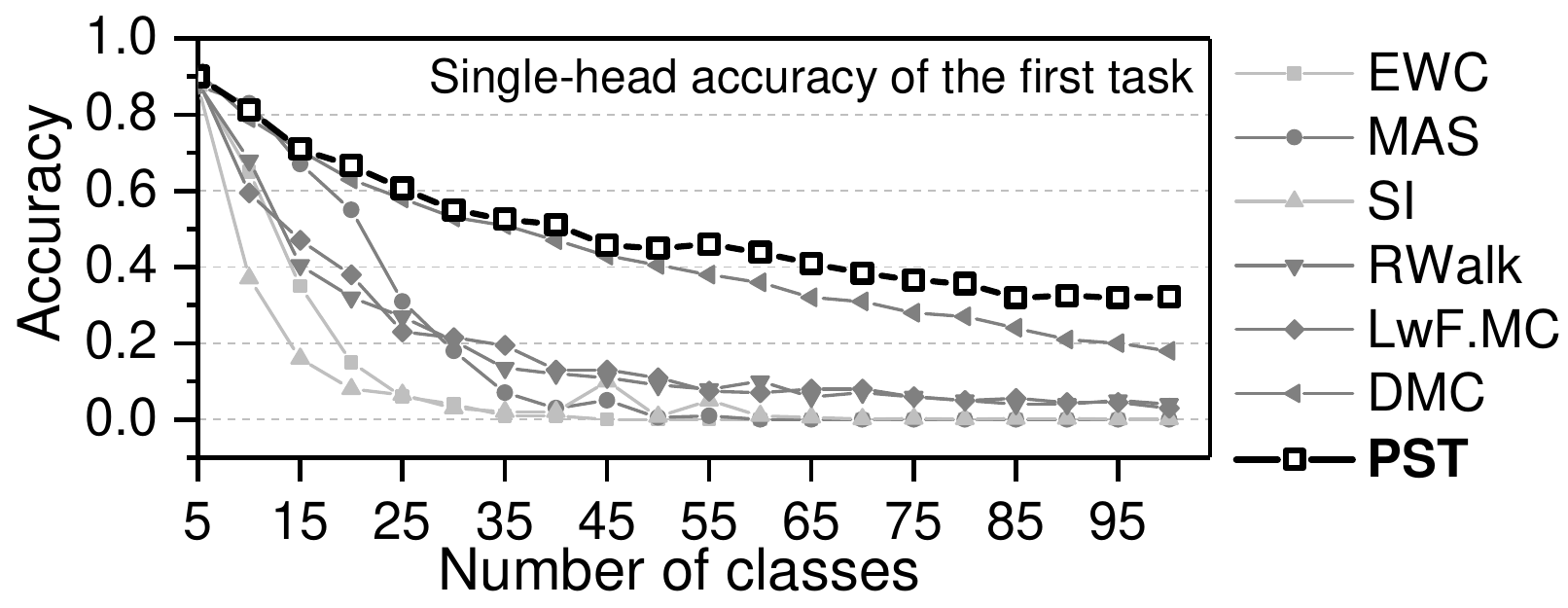}
\label{fig:single-head_task0}
}
\quad
\subfigure[Multi-head accuracy]{
\includegraphics[width=0.95\columnwidth]{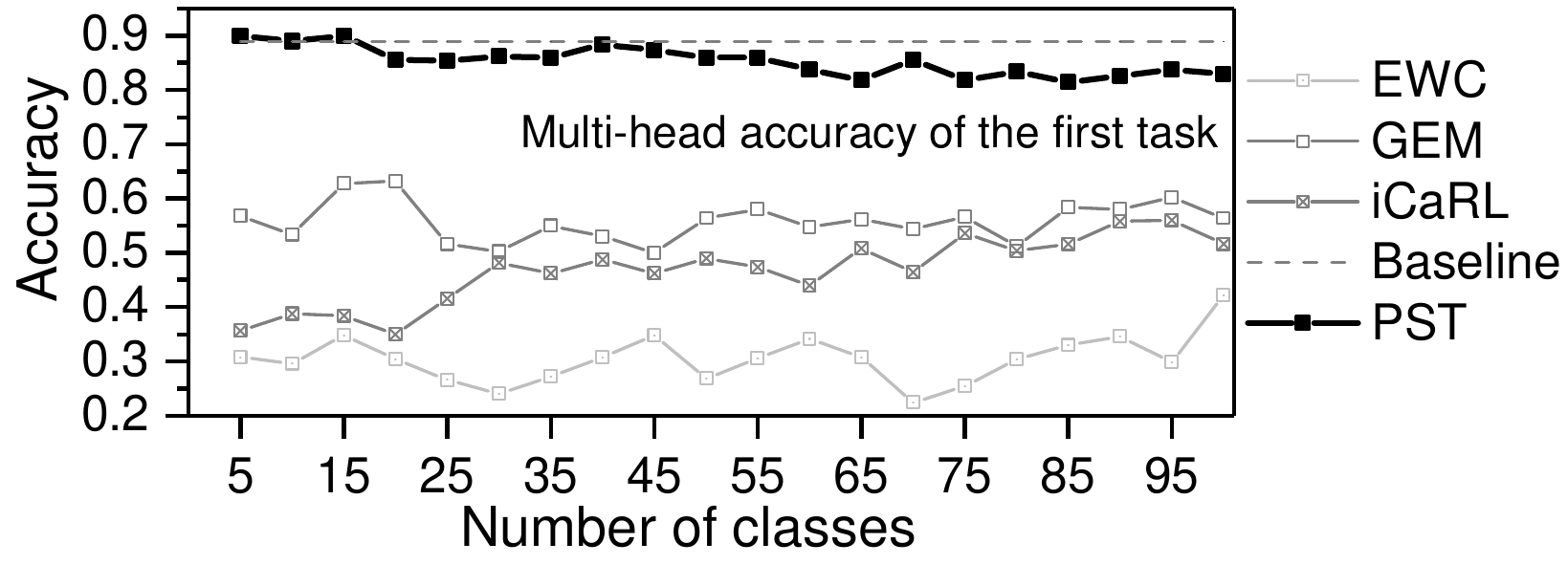}
\label{fig:multi-head_task0}
}
\caption{(a) Single-head accuracy and (b) multi-head accuracy of the first task $T_1$ over time when the model is trained with a sequence of 20 tasks on CIFAR-100.}
\label{fig:first_task}
\end{figure}

\textbf{Evaluation protocol.} As mentioned in Section~\ref{sec:intro}, single-head evaluation is more practical and valuable than multi-head evaluation in the scenario of continual learning Therefore, we evaluate single-head accuracy for the following experiments.  To report the single-head \textit{overall accuracy} if input data $\left\{X^{1}, \ldots, X^{t}\right\}$ have been observed so far, we test the network with testing data that sampled uniformly and randomly from class $1$ to class $t$ and predict a label out of $t$ classes $\left\{1, \dots, t\right\}$. For the \textit{first task accuracy} (such as Fig~\ref{fig:first_task}) , we test the network with testing data collected from the first task $T_1$ (supposing classes $\left\{1, \dots, g\right\}$) and predict a label out of $t$ classes $\left\{1, \dots, t\right\}$ to report single-head $T_1$ accuracy (Fig.~\ref{fig:single-head_task0}); or, predict a label out of $g$ classes $\left\{1, \dots, g\right\}$ to report multi-head $T_1$ accuracy (Fig.~\ref{fig:multi-head_task0}). 
\begin{figure}[!t]
\includegraphics[width=\columnwidth]{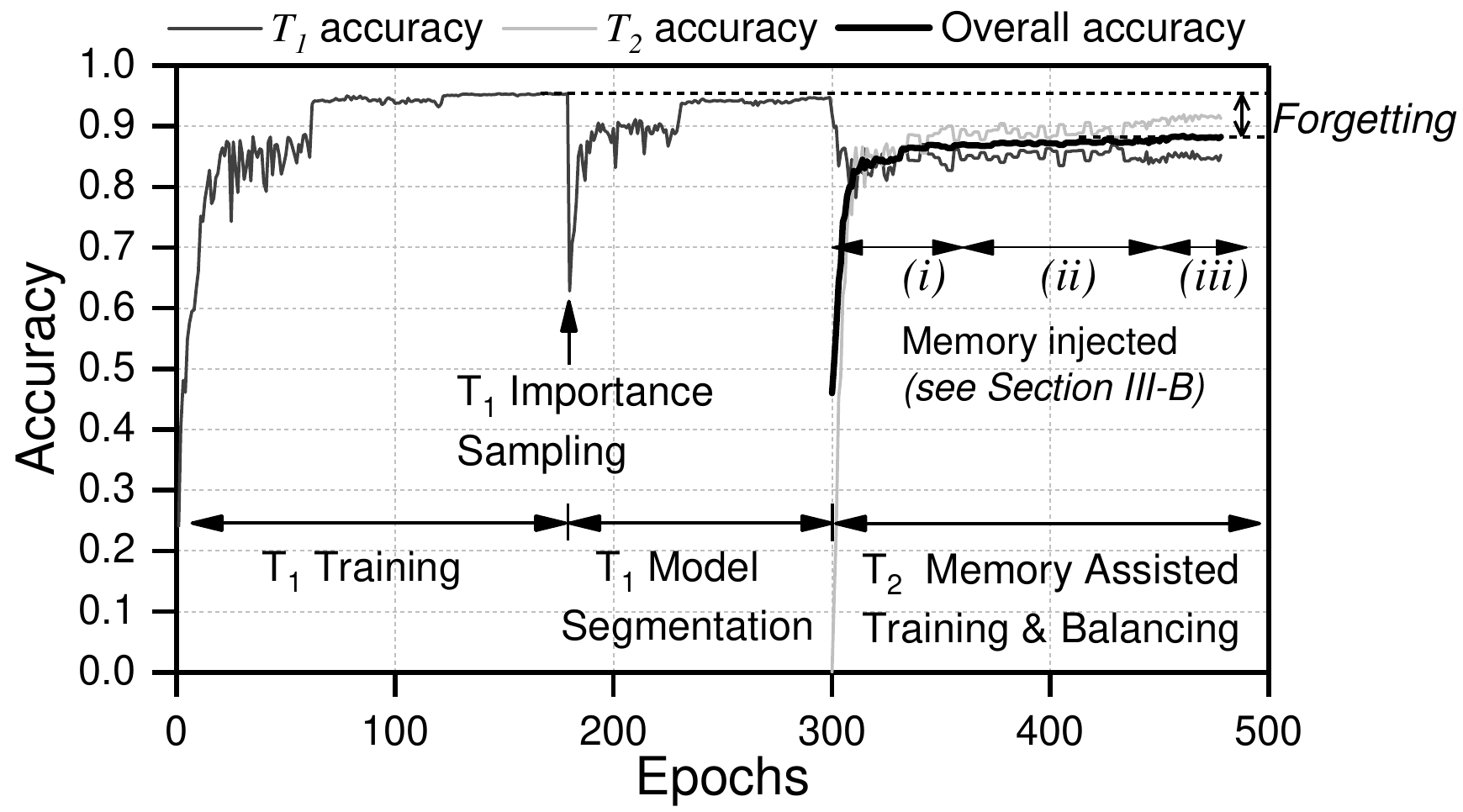}
\caption{The learning curve of 2 tasks on CIFAR-10 with each step annotated.}
\label{fig:LC}
\end{figure}
\begin{figure*}[!t]
\begin{center}
\includegraphics[width=\textwidth]{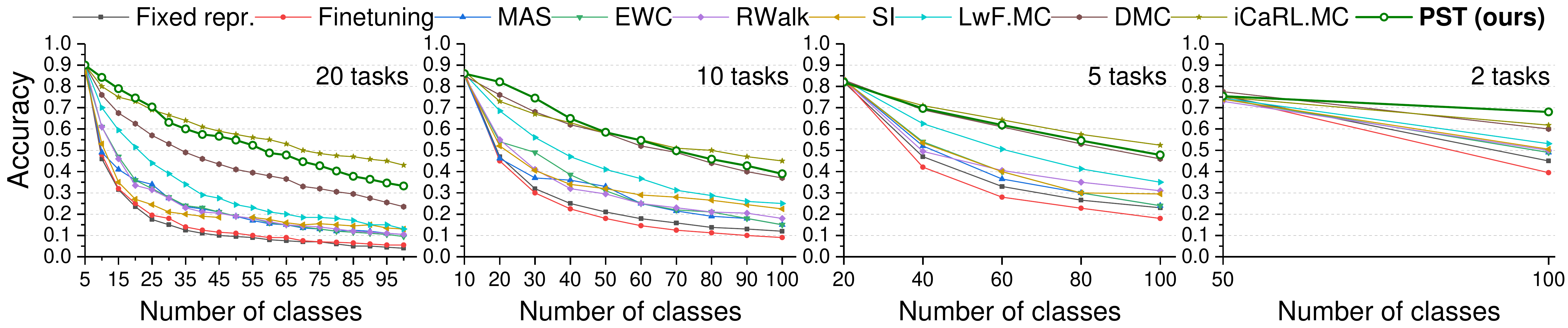}
\end{center}
\caption{Single-head overall accuracy on CIFAR-100 when incrementally learning 20, 10, 5, 2 tasks in a sequence. PST has the best accuracy of 2 tasks and the second best accuracy of 5, 10, 20 tasks. Though iCaRL.MC has better accuracy than PST, it requires $>$24$\times$ computation cost than PST (see Fig.~\ref{fig:flop} for details).}
\label{fig:overall}
\end{figure*}

\subsection{In-depth analysis}\label{sec:cifar10}

We divide CIFAR-10 into 2 tasks (5 classes each) and analyze the PST training routine step by step in this subsection.  The learning curve is present in Fig.~\ref{fig:LC}.

From epoch 0 to epoch 180, $T_1$ is trained and reaches baseline accuracy. The weight distribution after training $T_1$ is present in Fig.~\ref{fig:distribution}a.
At epoch 180, we sample the top 50\% (since there are 2 tasks totally) important parameters and retrain them with the secondary parameters untouched (epoch 180 to epoch 300), which is the model segmentation step.  
The weight distribution after this step is shown in  Fig.~\ref{fig:distribution}d. It is worth mentioning that previous works, such as PackNet~\cite{mallya_packnet:_2018} and Piggyback~\cite{ferrari_piggyback_2018}, prune the secondary parameters and thus, distort the weight distribution (Fig.~\ref{fig:distribution}b). 
At epoch 300, task $T_2$ appears and updates the parameters.

At the same time, the acquired knowledge of $T_1$ is disturbed by $T_2$ updating, leading to an accuracy degradation on $T_1$ (see the green curve at epoch 300). 
From epoch 300 to the end is the step of $T_2$ training, during which the memory data is injected following\textit{ (\romannumeral1), (\romannumeral2), and (\romannumeral3)} to balance. After $T_2$ training, we again plot the weight distribution for the pruning-based approach (in Fig.~\ref{fig:distribution}c) and PST approach (in Fig.~\ref{fig:distribution}e). It is observed that the pruning approach fails to preserve the prior knowledge, as the weight distribution after learning $T_2$ shifts far away from the previous one. In contrast, PST well preserves prior knowledge (\ie~similar weight distribution after learning $T_1$ and after learning $T_2$). Compared to the baseline accuracy, pruning-based approaches forget 31\% on overall accuracy while segmentation-based PST only forgets 5\%.

\subsection{Extensive results}
\label{sec:cifar100}

\textbf{Accuracy for incrementally learning multi-classes.}
We compare PST with state-of-the-art approaches that reported single-head accuracy: MAS\cite{ferrari_memory_2018}, EWC~\cite{kirkpatrick2017overcoming}, RWalk~\cite{chaudhry_riemannian_2018}, SI~\cite{zenke_continual_2017}, LwF.MC~\cite{li_learning_2018}, DMC~\cite{zhang_class-incremental_2019}, iCaRL.MC~\cite{rebuffi_icarl_2016} and two baselines: \textit{fixed representation}, \textit{finetuning}.\textit{ Fixed representation} denotes the method that we fix the feature extraction layers for the previously learned tasks and only train classification layers for new tasks. \textit{Finetuning} denotes the method that the network trained on previous tasks is directly fine-tuned by new tasks, without strategies to prevent catastrophic forgetting. LwF.MC denotes the method that uses LwF~\cite{li_learning_2018} but is evaluated with multi-class single-head classification. iCaRL.MC denotes the method uses iCaRL but replaces their Nearest-Mean-of-Exemplar~\cite{rebuffi_icarl_2016} classifier with a regular output classifier for a fair comparison with PST.   The results of MAS, EWC, RWalk, SI and DMC are from ~\cite{zhang_class-incremental_2019}, which is implemented with the official code\footnote{https://github.com/facebookresearch/agem}. The results of \textit{fix representation}, \textit{finetune}, LwF.MC and iCaRL are from~\cite{rebuffi_icarl_2016}. We adopt the same memory size for fair comparison between baselines and PST.

\begin{figure}[!t]
\centering
\includegraphics[width=0.85\columnwidth]{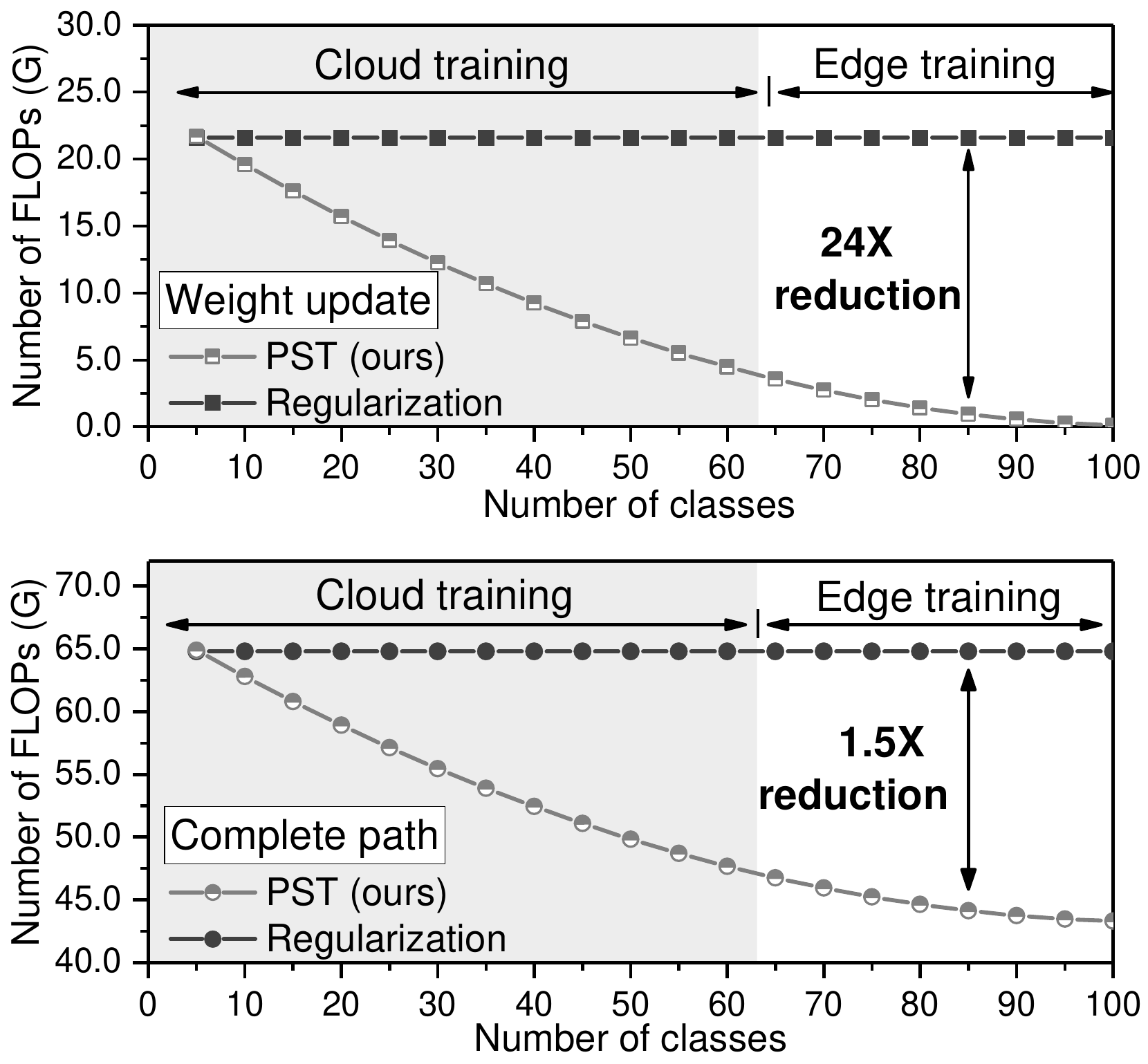}
\caption{Comparison of the computation cost of PST and the regularization method. In the scenario of edge learning, more than 24$\times$ reduction in FLOPs for the weight update path (top), and 1.5$\times$ reduction for complete path (bottom) are achieved.}
\label{fig:flop}
\end{figure}

The single-head overall accuracy when incrementally learning 20 tasks (5 classes per task), 10 tasks (10 classes per task), 5 tasks (20 classes per task) and 2 tasks (50 classes per task) are reported in Fig.~\ref{fig:overall}. Among 9 different approaches, PST achieves the best accuracy on the 2-task scenario and the second best accuracy on the other scenarios. Compare to \textit{finetuning}, PST largely prevents the model from catastrophic forgetting.  Though PST achieves lower accuracy than iCaRL in some cases, PST is more than 24$\times$ efficient in computation cost,  as shown in Fig.~\ref{fig:flop}. This efficiency is benefiting from model segmentation: iCaRL has to update the entire network parameters for every new observation, but PST only requires updating partial network parameters, as the parameters related to previous tasks are frozen. Meanwhile, PST outperforms iCaRL in the multi-head protocol, as present in the following paragraph.

\textbf{Accuracy of the first task.}
Fig.~\ref{fig:single-head_task0} compares the single-head accuracy on the first task $T_1$ in PST with several previous approaches that reported $T_1$ accuracy in their papers. It also presents the multi-head accuracy on $T_1$ in PST and the baseline accuracy. PST achieves the best single-head accuracy on $T_1$ among all the approaches, \ie~the least forgetting. Moreover, when $T_1$ data is evaluated in a multi-head classification setting, as shown in Fig.~\ref{fig:multi-head_task0}, PST is stable and always on par with the baseline (the model that is only trained on $T_1$, so without forgetting). This phenomenon demonstrates that PST effectively preserves the knowledge related to $T_1$ through model segmentation. 
Without these strategies, it is hard to keep the previously acquired knowledge. For example, GEM~\cite{lopez-paz_gradient_2017} reported unstable multi-head $T_1$ accuracy, which is because the parameters gradually drift away from $T_1$ knowledge after a long period of learning on new tasks.


\subsection{Efficient edge learning}
\label{sec:edge}

In a more realistic situation (such as self-driving cars), continual learning may not be used to train a model from scratch at the edge.  Instead, we will have a model which is well trained in the cloud (\ie~knowledge repository) and once deployed, might only be required to learn \textit{a few} new classes in an online manner on the edge devices. For example, 90 classes of CIFAR-100 are pre-trained in the cloud and 10 classes are expected to be learned on edge devices continually. Moreover, computation cost is a critical overhead when deploying Deep Neural Networks on edge devices~\cite{han2015learning, haoli, luo2017thinet, 8755608}. Therefore, edge learning prefers algorithms with low computation cost rather than that with higher one.


For such a scenario, we estimate the training computation cost, \ie~the number of floating point operations (FLOPs), required by PST and regularization approaches such as iCaRL~\cite{rebuffi_icarl_2016} and EWC~\cite{kirkpatrick2017overcoming}, as shown in Fig.~\ref{fig:flop}.  Training on hardware includes three paths \cite{ko2019automatic}, \ie~(1) forward path, (2) backward path, and (3) weight update path. 
As more and more tasks come in, the trainable parameters become fewer and fewer in PST, \ie, weight update path gradually requires fewer operations, but regularization methods require a constant number of operations at all times, as the model is not segmented. Thus, given the model is pre-trained in the cloud with a large amount of data (\eg~90 classes) and loaded at the edge to learn new observations (\eg~10 classes), PST reduces more than 24$\times$ FLOPs in the weight update path, and more than 1.5$\times$ FLOPs in complete path (including all three paths), as compared to the regularization methods such as iCaRL~\cite{rebuffi_icarl_2016}. Especially, weight update path usually costs 2$\times$ latency than the other two paths so that PST can largely speed up the training. 
Benefiting from segmentation, PST outperforms other continual learning schemes in computation efficiency.


\begin{figure}[!t]
\centering
\includegraphics[width=0.75\columnwidth]{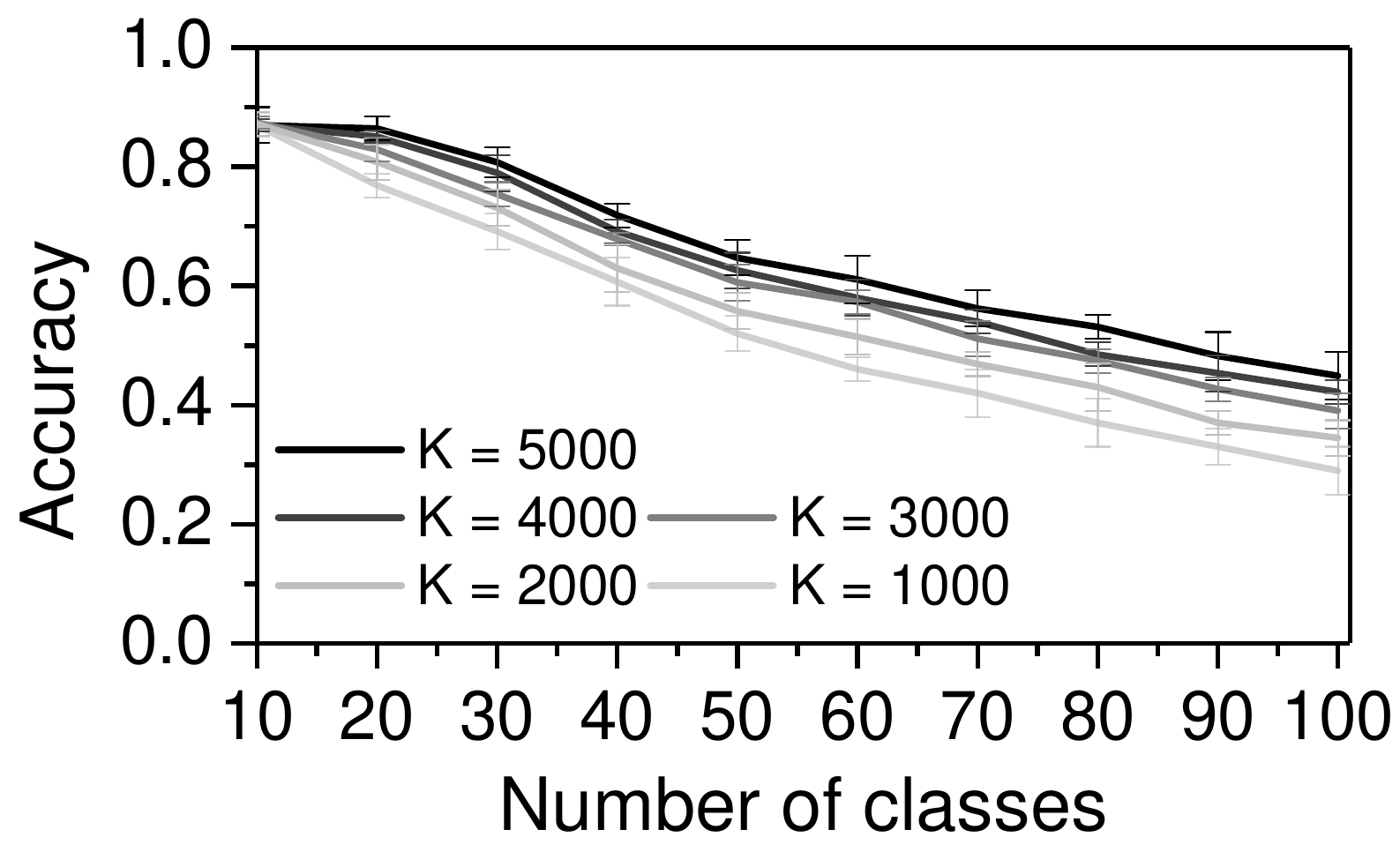}
\caption{Overall single-head accuracy when incrementally learning 10 tasks under different memory budget.}
\label{fig:memory}
\end{figure}

\begin{table}[!t]
\caption{Switching off different components of PST leads to accuracy drop to different extents. In the table, negative numbers indicate accuracy drop,~\eg~-0.32 means 32\% accuracy drop. }
\label{table:hybrid}
\centering
\resizebox{\columnwidth}{!}{ 
\begin{tabular}{|c|c|c|c|}
 \hline
\textbf{Model}  &      \textbf{20 tasks} & \textbf{10 tasks} & \textbf{5 tasks}  \\ \hline
Hybrid 1 (removing importance sampling)   &  -0.32      & -0.38     & -0.45  \\\hline
Hybrid 2 (removing model segmentation)   &  -0.32      & -0.38     & -0.42   \\\hline
Hybrid 3 (removing memory balancing) &  -0.06      & -0.08     & -0.11    \\\hline
\end{tabular}
}
\end{table}

\section{Ablation study and discussion}\label{sec:discussion}
In this section, we analyze the importance of each component in PST by performing an ablation study and demonstrate that each component in PST is contributing to the overall performance. 

Fig.~\ref{fig:memory} illustrates the effect of different memory budgets. With more data saved from previous tasks, the forgetting is reduced. But such a trend gradually saturates. Considering resource constraint, the learning system cannot rely on memory rehearsal itself to recover catastrophic forgetting. 

To analyze the role of each component in PST, we switch off each component in PST and repeat the experiments performed in Fig.~\ref{fig:overall}.
 Replacing importance sampling with a random sampling leads to model \textit{Hybrid 1}; removing model segmentation step (no reinforcement on $\Theta_{important}$) leads to model \textit{Hybrid 2}; removing the memory-assisted balancing  leads to model \textit{Hybrid 3}. 
 The overall accuracy \textit{change} after the last task is reported in Table~\ref{table:hybrid}.
The results of hybrid models prove that importance sampling and model segmentation are indispensable steps for PST since removing them leads to significant accuracy drop, and memory-assisted balancing is supplementary.


\section{Conclusion}\label{sec:conclusion}

A successful continual learning system that is exposed to a continuous data stream should have the properties of online adaption, preservation of prior knowledge, single-head evaluation and resource constraint, to alleviate or even prevent catastrophic forgetting of previously acquired knowledge. To satisfy these properties and minimize catastrophic forgetting, we propose a novel scheme named single-net continual learning with Progressive Segmented Training (PST). Benefiting from its components (memory-assisted training and balancing, importance sampling, and model segmentation), PST achieves state-of-the-art single-head accuracy on incremental tasks on CIFAR datasets, with far lower computation cost. We further demonstrate that PST favors edge computing due to its segmented training method. In future work, we plan to  study the detailed mechanism of catastrophic forgetting further and improve PST. Moreover, we plan to explore compressing or even eliminating the memory data without sacrificing the performance.

\section*{Acknowledgment}
This work was supported in part by C-BRIC, one of six centers in JUMP, a Semiconductor Research Corporation (SRC) program sponsored by DARPA. It is also partially support by National Science Foundation (CCF \#1715443).

\bibliographystyle{IEEEtran.bst}
\bibliography{pst}

\end{document}